\documentclass[runningheads]{llncs}
\usepackage{graphicx}
\usepackage{hyperref}
\usepackage{fancyvrb}

\usepackage{xcolor}
\usepackage{subcaption}

\usepackage{color, colortbl}
\definecolor{Gray}{gray}{0.9}

\begin{document}
\title{Certified Control: An Architecture for Verifiable Safety of Autonomous Vehicles}

\titlerunning{Certified Control}
%
\author{
    \mbox{Daniel Jackson\inst{1}} \and
    \mbox{Valerie Richmond\inst{1}} \and
    \mbox{Mike Wang\inst{1}} \and
    \mbox{Jeff Chow\inst{1}} \and
    \mbox{Uriel Guajardo\inst{1}} \and
    \mbox{Soonho Kong\inst{2}} \and
    \mbox{Sergio Campos\inst{3}}  \and
    \mbox{Geoffrey Litt\inst{1}} \and
    \mbox{Nikos Arechiga\inst{2}}
}
%
\authorrunning{D. Jackson et al.}
%
\institute{
  MIT \and
  Toyota Research Institute \and
  Universidade Federal de Minas Gerais, Brazil
}

\maketitle
\begin{abstract}
Widespread adoption of autonomous cars will require greater confidence in their safety than is currently possible. Certified control is a new safety architecture whose goal is two-fold: to achieve a very high level of safety, and to provide a framework for justifiable confidence in that safety. The key idea is a runtime monitor that acts, along with sensor hardware and low-level control and actuators, as a small trusted base, ensuring the safety of the system as a whole.

Unfortunately, in current systems complex perception makes the verification even of a runtime monitor challenging. Unlike traditional runtime monitoring, therefore, a certified control monitor does not perform perception and analysis itself. Instead, the main controller assembles evidence that the proposed action is safe into a \textit{certificate} that is then checked independently by the monitor. This exploits the classic gap between the costs of finding and checking. The controller is assigned the task of finding the certificate, and can thus use the most sophisticated algorithms available (including learning-enabled software); the monitor is assigned only the task of checking, and can thus run quickly and be smaller and formally verifiable.

This paper explains the key ideas of certified control and illustrates them with a certificate for LiDAR data and its formal verification. It shows how the architecture dramatically reduces the amount of code to be verified, providing an end-to-end safety analysis that would likely not be achievable in a traditional architecture.

\end{abstract}


\section{Introduction}
If autonomous cars are to become widespread, it will be
necessary not only to ensure a high level of safety but also to
justify our confidence that such a level has been achieved.

Formal verification seems essential, but cars typically make decisions using machine learning and complex algorithms, which are hard to verify. Rather than attempting to extend the range of verification techniques, this paper proposes \textit{certified control}, a new architecture that allows existing techniques to be applied in this challenging setting.

The key idea is a variant on the traditional concept of the runtime monitor. Like a conventional monitor, the monitor checks actions proposed by the main controller before passing them on to the actuators; if an action is found to be unsafe, it is blocked or replaced by a safer action.

Unlike a conventional monitor, however, the monitor does not make this decision based on its own interpretation of the environment. Instead, it checks a \textit{certificate} generated by the main controller that contains, in addition to the proposed action, explicit evidence that the action is safe. When the monitor approves the certificate, it has essentially ratified a runtime safety case.

The generation of the certificate is the responsibility of the main controller, and is a byproduct of its normal analysis. The form of the certificate, and the argument that a successful check ensures safety, are developed at design time.


Certified control exploits four powerful computer science
ideas in a novel combination and context: (1) the gap between
the complexities of \textbf{finding vs. checking}; (2) the idea of a
\textbf{small trusted base} (here, the monitor and sensors and actuators) allowing a guarantee of safety without dependence on unreliable components outside the base (in particular the perception system); (3) the use of \textbf{authentication} (e.g., of sensor data) to
guarantee safe transmission through an untrusted channel
(namely the main perception/controller subsystems); and (4)
the idea of \textbf{end-to-end safety cases} whose form is justified at design time, but which are applied on the fly. 

The consequence of this design is that some essential safety guarantees that might have required a verification of the entire system can now be established with a verification of a much smaller part.

The contributions of this paper include: (1) A review of existing approaches to ensuring safety in critical systems; (2) A characterization of the demands on a runtime monitor in terms of three essential properties; (3) An argument that, for autonomous vehicles, the classic runtime monitor (or ``safety
controller" approach) is unlikely to satisfy all three properties, and is thus not ideal; (4) A new architecture, called \textit{certified control}, that achieves the three properties; (5) An evaluation of the approach in an application to LiDAR-based obstacle detection, with empirical validation through a race car implementation and a formal verification; (6) An analysis of the potential and limitations of the approach.

 The approach is also being applied to other forms of perception; the appendix describes certificates for lane-following that allow a similar reduction in complexity but have yet to be formally verified.



\section{Background}
The problem of safety for self-driving cars has two distinct
aspects. First is the reality of numerous accidents, many fatal,
either involving fully autonomous cars---such as the Uber that
killed a pedestrian in Tempe, Arizona~\cite{wakabayashiSelfDrivingUberCar2018}---or cars with
autonomous modes---such as the Tesla models, which have
spawned a rash of social media postings in which owners have
demonstrated the propensity of their own cars to repeat
mistakes that had resulted in fatal accidents. The metric of
``miles between disengagements,'' made public for many
companies by the California DMV~\cite{TestingAutonomousVehicles}, has revealed the
troublingly small distance that autonomous cars are apparently able to travel without human intervention. Even if the disengagement metric is crude and includes disengagements that are not safety-related~\cite{cameronDriverlessReadinessScore}, the evidence suggests that the technology still has far to go.

Second, and distinct from the actual level of safety
achieved, is the question of confidence. Our society's
willingness to adopt any new technology relies on our
confidence that catastrophic failures are unlikely. But, even
for the designs with the best records of safety to date, the
number of miles traveled falls far short of the distance that
would be required to provide statistical confidence of a failure
rate that matches (or improves on) the failure rate of an
unimpaired human driver. Even though Waymo, for example,
claims to have covered 20 million miles—a truly impressive
achievement—this still pales in comparison to the 275 million
miles that would have to be driven for a 95\% confidence that
fully autonomous vehicles have a fatality rate lower than a
human-driven car (one in 100 million miles)~\cite{kalraDrivingSafetyHow2016}. In fact, the problem is even worse, because the statistical evidence of few accidents given miles driven will be invalidated as soon as changes are made to the code.\footnote{Thank you to Ben Sherman for this astute and important observation.}

\subsection{An alternative to testing}
Statistical testing is the gold standard for quality control for many products (such as pharmaceuticals) because it is
independent of the process of design and development. This
independence is also its greatest weakness, because it denies the designer the opportunity to use the structure of the artifact
to bolster the safety claim, and at the same time fails to focus testing on the weakest points of the design, thus reducing
the potency of testing for establishing near-zero likelihood of
catastrophic outcomes.

One alternative to statistical testing is to construct a ``safety
case,'' an argument for safety based on the structure of the
design~\cite{weinstockDependabilityCases2004}. The quality of the argument and the extent to which
experts are convinced then becomes the measure of
confidence. This approach lacks the scientific basis of
statistical testing, but is widely accepted in all areas of
engineering, especially when the goal is to prevent catastrophe
rather than a wider range of routine failures. For example,
confidence that a new skyscraper will not fall down relies not
on testing (since each design is unique, and non-destructive
tests reveal little) but on analytical arguments for stability and
resilience in the presence of anticipated forces. In the UK, the
use of safety cases is mandated by a government standard~\cite{SafetyManagementRequirements2007} for critical systems such as
nuclear power plants.

In software too, there is growing interest in safety cases (or,
more generally, assurance or dependability cases)~\cite{jacksonSoftwareDependableSystems2007}.
For a cyber-physical system, the safety case is an
argument that a machine, in the context of its environment,
meets certain critical requirements. This argument is a chain of
many links, including: the specification of the software that
controls the machine, the physical properties of the
environment (including peripheral devices such as sensors and
actuators that mediate between the machine and the
environment), and assumptions about the behavior of human
users and operators. Each link in the chain needs its own
justification, and together they must imply the requirements.
Ideally, the justification takes the form of a mathematical
proof: in the case of software, for example, a verification
proof that the code meets the specification. But some links
will not be amenable to mathematical reasoning: properties of
the environment, and of physical peripherals, for example, must be formulated and justified by expert inspection.

\subsection{The Cost of Verification}
For software-intensive systems, the software itself can become
a problematic link in the chain. Complex systems require
complex software, and that inevitably leads to subtle bugs.
Because the state space of a software system is so large,
statistical testing can only cover a tiny portion of the space,
and thus cannot provide confidence in its correctness. So for
high confidence, verification seems to be the only option.

Unfortunately, verification is prohibitively expensive. Even
for software produced under a very rigorous process that does
not involve verification, the cost tends to be orders of
magnitude higher than for conventional software development.
NASA's flight software, for example, has cost over \$1,000 per
line of code, where conventional software might cost \$10 to
\$50 per line~\cite{TheyWriteRight}. Verifying a large codebase is a Herculean task. It
may not be impossible, as demonstrated by the success of
recent projects to verify an entire operating system kernel or file
system stack. But it typically requires enormous manual
effort. SEL4, a verified microkernel, for example, comprised
about 10,000 lines of code, but required about 200,000 lines of hand-authored proof, whose production took about 25-30
person years of work~\cite{kleinSeL4FormalVerification2009}.

For autonomous vehicles, the extensive use of machine learning makes it questionable whether verification is feasible at any cost (see Section \ref{sec:related} for some recent work on verifying neural networks).

\subsection{Small trusted bases}
One way to alleviate the cost of verification is to design the software system so that it has a small trusted base. The trusted base is the portion of the code on which the critical safety properties depend; any part of the system outside the trusted base can fail without compromising safety. This idea is
exploited, for example, in secure transmission protocols that employ encryption (and is generalized in the ``end-to-end principle''~\cite{saltzerEndtoendArgumentsSystem1984}). So long as the encryption and decryption
algorithms that execute at the endpoints are correct, one can be
sure that message contents are not corrupted or leaked; the
network components that handle the actual transmission, in
particular, need not be secure, because any component that
lacks access to the appropriate cryptographic keys cannot
expose the contents of messages or modify them without the
alteration being detectable.

Of course, the claim that some subset of the components of
a system form a trusted base---really that the other
components fall outside the trusted base---must itself be
justified in the safety case. It must be shown not only that the properties established by the trusted base are sufficient to
ensure the desired end-to-end safety properties, but also that the trusted base is immune to external interference that might cause it to fail (a property often achieved by using separation mechanisms to isolate the trusted base).

\subsection{Runtime monitors and safety controllers}
One widely-used approach is to augment the system with a
runtime monitor that checks (and enforces) a critical safety
property. If isolated appropriately, and if the check is
sufficient to ensure safety, the monitor serves as a trusted
base.


For safety-critical systems, the runtime monitor might be an
entire controller in its own right. This ``safety controller''
oversees the behavior of the main controller, and takes over
when it fails. If the safety controller is simpler than the main
controller, it serves as a small trusted base (along with
whatever arbiter is used to ensure that it can veto the main
controller's outputs). This scheme is used in the Boeing 777,
which runs a complex controller that can deliver highly
optimized behavior over a wide range of conditions, but at the
same time runs a secondary controller based on the control
laws of the 747, ensuring that the aircraft flies within the
envelope of the earlier (and simpler) design~\cite{yehDependability777Primary1995}.

The Simplex architecture ~(\cite{crenshawSimplexReferenceModel2007,luishaUsingSimplicityControl2001}) embodies this idea in a general
form. Two control subsystems are run in parallel. The high
assurance subsystem is meticulously developed with
conservative technologies; the high performance subsystem
may be more complex, and can use technologies that are hard
to verify (such as neural nets). The designer identifies a safe region of states that are within the operating constraints of the system and which exclude unsafe outcomes (such as collisions). A smaller subset of these states, known as the \textit{recovery} region is
then defined as those states from which the high assurance
subsystem can always recover control and remain within the
safe region. The boundary of the recovery region is then used as the switching condition between the two subsystems.

\subsection{The problem of perception in autonomous cars}
The safety controller approach relies on the assumption that
the controller itself is the most complex part of the
system—that from the safety case point of view, the
correctness of the controller is the weakest link in the
argument chain. But in the context of autonomous cars,
perception---the interpretation of sensor data---is more
complicated and error-prone than control. In particular,
determining the layout of the road and the presence of
obstacles typically uses vision systems that employ large and
unverified neural nets.

In standard safety controller architectures (such as
Simplex~\cite{crenshawSimplexReferenceModel2007,luishaUsingSimplicityControl2001}),
only the controller itself has a safety counterpart; even if sensors
are replicated to exploit some hardware redundancy, the conversion of
raw sensor data into controller inputs is performed externally to the
safety controller, and thus belongs to the trusted base.

This means that the safety case must include a convincing
argument that this conversion, performed by the perception
subsystem, is performed correctly. This is a formidable task
for at least two reasons. First, there is no clear specification
against which to verify the implementation. Machine learning
is used for perception precisely because no succinct, explicit articulation of the expected input/output relationship is readily
available. Second, state of the art verification technology
cannot handle the particular complications of deep neural networks---especially
their scale and use of non-linear activation functions (such as
ReLU~\cite{nairRectifiedLinearUnits2010}) which confound automated
reasoning algorithms such as SMT and linear programming~\cite{pulinaChallengingSMTSolvers2012}.

An alternative possibility is to not include perception
functions in the trusted base. Instead, one could perhaps use a
runtime monitor that incorporates both a safety controller and
a simplified perception subsystem. Initially, this approach
seemed attractive to us, but we came to the conclusion that it
was not in fact viable. In the next section we explain why.

\subsection{Desiderata for a runtime monitor: choose two}
To see why a runtime monitor that employs simplified
perception is not a solution to the safety problem for autonomous cars, we shall enumerate three critical properties that a monitor should obey, and argue that---in this context--they are mutually inconsistent for the conventional design.

The first property is that the monitor should be \textbf{verifiable}.
That is, it should be small and simple enough to be amenable
to formal verification (or to exhaustive testing).
If not, the monitor brings no significant benefit in terms of confidence in the overall system safety (beyond the diversity of an additional implementation, which contributes less confidence than is often assumed~\cite{luishaUsingSimplicityControl2001}).

The second property is that the monitor should be \textbf{tolerant}. It
should intervene only when necessary, namely when
proceeding with the action proposed by the main controller
would be a safety risk. Applying emergency braking on a
highway when there is no obstacle, for example, is clearly
unacceptable. Even handing over control to a human driver is
problematic, due to vigilance decrement~\cite{greenleeDriverVigilanceAutomated2018}.

The third property is that the monitor should be \textbf{sensitive}.
This means that it should ensure the safety of the vehicle
within an envelope that covers a wide range of typical
conditions. It is not sufficient, for example, for the monitor to
merely reduce the severity of a collision when it might have
been able to avert the collision entirely.

Unfortunately, it seems that these three properties cannot be
achieved simultaneously in a classic monitor design. The
problem, in short, is that the combination of tolerance and
sensitivity requires a sophisticated perception system,
leading to unverifiable complexity. It is easy to make a
monitor that is sensitive but not tolerant simply by not allowing
the vehicle to move; and conversely it is easy to make one that
is tolerant but not sensitive by not preventing any collisions at
all. Achieving both at once, however, seems inevitably to require a complex and 
potentially unverifiable perception component.

As a sneak preview of an example discussed in more detail later, consider a monitor that detects obstacles with LiDAR, filtering out nearby reflections of snow by using an algorithm to distinguish snow flakes from real obstacles. Using such an algorithm challenges verifiability; not using it likely either violates tolerance (if snowflakes are treated as obstacles) or sensitivity (if a cruder algorithm is used that filters out not only snowflakes but also genuine obstacles). For monitors that involve vision (as explained in the appendix), the problem is exacerbated yet further, since neural nets are generally required.

This does \textit{not} mean that monitors that fail to satisfy these three properties are not useful---only that such monitors are not \textit{sufficient} to ensure safety. Automatic emergency braking
(AEB) systems are deployed in many cars now, and use
radar to determine when a car in front is so close that braking is essential. But because AEB might brake too late to prevent a collision, it is not generally sensitive. Responsibility-Sensitive Safety systems~\cite{shalev-shwartzFormalModelSafe2018}, on the other hand, use the car's full sensory perception systems to identify and locate other cars and pedestrians, and can therefore be both tolerant and sensitive, but due to the complexity of the
perception are not likely to be verifiable.


\section{Certified Control: A New Approach}


Certified control (see Fig.~\ref{fig:certified_control_diagrams} in Appendix~\ref{architecture-appendix}) is a new architecture that centers on a
different kind of monitor. As with a conventional safety
architecture, a monitor vets
proposed actions emanating from the main controller. But the
certified control monitor does not check actions against its
\textit{own} perception of the environment.

Instead, it relies on the main perception and control
subsystems to provide a \textit{certificate} that embodies evidence
that the situation is safe for the action at hand. The certificate
is designed to be unforgeable, so that even a malicious agent
could not convince the monitor that an unsafe situation is safe.
The evidence comprises sensor readings that have been
selected to support a safety case in favor of the proposed
action; because these sensor readings are only \textit{selected} by the main perception and control subsystems (and are signed by the sensor units that produced them), they cannot
be faked.

A certificate contains the following elements: (1) the
proposed action (for example, driving ahead at the current
speed); (2) some signed sensor data (for example, a set of
LiDAR points or a camera image); (3) optionally, some
interpretive data. This data indicates what inference should be
drawn from the sensor readings. For example, if the sensor
data comprises LiDAR points intended as evidence that the
nearest obstacle is at least some distance away, the interpretive
data might be that distance; if the sensor data is an image of
the road ahead, the interpretive data might be the purported lane
lines.

The evidence and interpretive data are evaluated by the
certificate checker using a predefined runtime safety case. As
an example, consider a certificate that proposes the action to
continue to drive ahead using LiDAR data. In this case, the
LiDAR points argue that there is no obstacle along the path;
each LiDAR reading provides direct physical evidence of an
uninterrupted line from the LiDAR unit to the point of
reflection. Together, a collection of such readings, covering
the cross section of the path ahead with appropriate density,
indicates absence of an obstacle larger than a certain size.

Compare this with a classic monitor that interprets the
LiDAR unit's output itself. The LiDAR point cloud is likely to
include points that should be filtered out. In snow, for
example, there will be reflections from snowflakes. Performing snow
filtering would introduce complexity and likely render the
monitor unverifiable. On the other hand, a simple monitor
would not attempt to identify snow, and could set a low or a
high bar on intervention—requiring, say, that 10\% or 90\% of
points in the LiDAR point cloud show reflections within some
critical distance. The low bar would result in a monitor that
violates sensitivity, failing, for example, to
prevent collision with a motorcycle whose cross section
occludes less than the 10\% of points. The high bar would
result in a monitor that violates tolerance, since
it would likely cause an intervention due to snow even when
the road ahead is empty of traffic.

In contrast, the certified control monitor does not suffer from these
problems, and can achieve verifiability, tolerance and sensitivity
together. It should be noted that certified control does not remove the
sensor and actuator units from the trusted base. What is
removed is the main perception and controller subsystems,
which crucially include complex algorithms that process and
interpret sensor readings.


\section{A Case Study}

We conducted a case study to evaluate the certified control architecture, with three research questions in mind:

\begin{enumerate}
    \item \textbf{RQ1}: Is it possible even to build a system based on certified control in the context of an autonomous car that has plausible behavior, demonstrating a reasonable combination of tolerance and sensitivity?
    \item \textbf{RQ2}: What reduction in the trusted base is achievable by following this strategy?
    \item \textbf{RQ3}: Is the resulting runtime monitor that checks certificates potentially verifiable?
\end{enumerate}

To answer these questions, we designed and implemented certificates for three separate monitors: (1) the example alluded to above, in which a certificate captures the result of LiDAR snow filtering; (2) a scheme for checking the result of a visual lane detection algorithm; (3) an augmented version of the lane detection example that combines both visual and LiDAR data to ensure that the purported lane lines lie on the ground plane. Only the first of these is discussed in detail in this paper; the other two are discussed at length in the appendix.

In each case, we ran tests using a 1/10th model racecar to explore the performance of the monitors. In the first case, we verified the code of the monitor and showed that it implies the desired safety property. We have not yet verified the implementations of the other two monitors, although we have measured the reduction in the size of the trusted base, which shows reassuring results.

\section{A LiDAR Certificate for Snow Filtering}\label{sec:lidar_cert}
Our primary example is a certificate based on LiDAR evidence that allows the car to continue ahead, with the certificate providing evidence that there is no obstacle within the car's stopping distance. 

Our controller implements radius outlier removal (ROR) filtering~\cite{charronDenoisingLidarPoint2018} to identify LiDAR points that correspond to reflections off snowflakes. The controller preprocesses the LiDAR point cloud into a k-dimensional tree to query for nearest neighbor information. Points with few neighbors relative to the average neighborhoods are labeled as snow. The controller then constructs the certificate omitting those snow points.

The certificate includes all the parameters needed to establish the safety of continuing ahead in lane. (Determining where the lane actually lies is a separate problem; see Appendix~\ref{lane-appendix} for details of a certificate scheme for that.) These parameters, along with the precise specification of the certificate check, are defined in Appendix~\ref{proof-appendix}. Informally, the parameters include:

\begin{itemize}
    \item An array of arrays of LiDAR points, correspond to horizontal rows;
    \item An array of row heights, corresponding roughly to the heights of the points in each row;
    \item Boundaries of a rectangle representing the cross section of the lane, specified as leftmost and rightmost positions on the x-axis (which runs left to right), and top and bottom positions on the y-axis (which runs top to bottom);
    \item The stopping distance at the current velocity\footnote{Our current implementation assumes the car is traveling at its maximum speed. We did this because our race car does not provide sufficiently accurate velocity measurements. Extending to arbitrary velocity is straightforward.};
    \item Constants representing maximum allowed spacing between points in the same row horizontally, between rows vertically, and vertically between the points in a row and the rows's purported height.
\end{itemize}

The monitor checks three key properties:
\begin{itemize}
\item \textit{Distance}: Every point is no closer than the minimum stopping distance.
\item \textit{Spread}: The first and last points within each row are within the maximum allowed horizontal spacing of the leftmost and rightmost lane rectangle boundaries, and the row heights likewise span from the top to the bottom;
\item \textit{Density}: Contiguous points\footnote{Since the LiDAR points represent reflections at different distances, they are first projected onto a perpendicular plane at the stopping distance.} within a row are within the maximum allowed horizontal spacing; contiguous elements in the row height array are within the maximum vertical spacing; and the points within a row deviate from the row's height by less than the maximum permitted amount.
\end{itemize}

The key elements of the safety argument are (a) the correctness of the monitor (namely that its code checks the certificate correctly); and (b) that the certificate properties imply the safety requirement. This latter argument is as follows: given the maximum spacing between points vertically (the sum of the maximum row spacing and height deviation of points within a row) and the maximum spacing of points horizontally, it follows that any obstacle taller and wider than those spacings cannot be closer than the stopping distance. This argument was formalized and checked using the SMT solver Z3, for LiDAR point arrows containing up to 100 points, and is detailed in Appendix~\ref{proof-appendix}.

\subsection{Experimental Evaluation}

The primary experimentation platform was a 1/10-scale remote-controlled car, outfitted with a Velodyne Puck VLP16 LiDAR scanner, a camera, and a CPU running Linux with ROS. Python2 implementations of a controller and interlock are loaded onto the car and executed in real time. We manipulated the car's environment to reflect various road conditions using props, like confetti for snow. The complete source code for our experimental implementations is publicly available; a URL will be provided after blind review.

\begin{figure}[!ht]
  \centering
  \begin{subfigure}[t]{0.24\linewidth}
    \includegraphics[width=\textwidth]{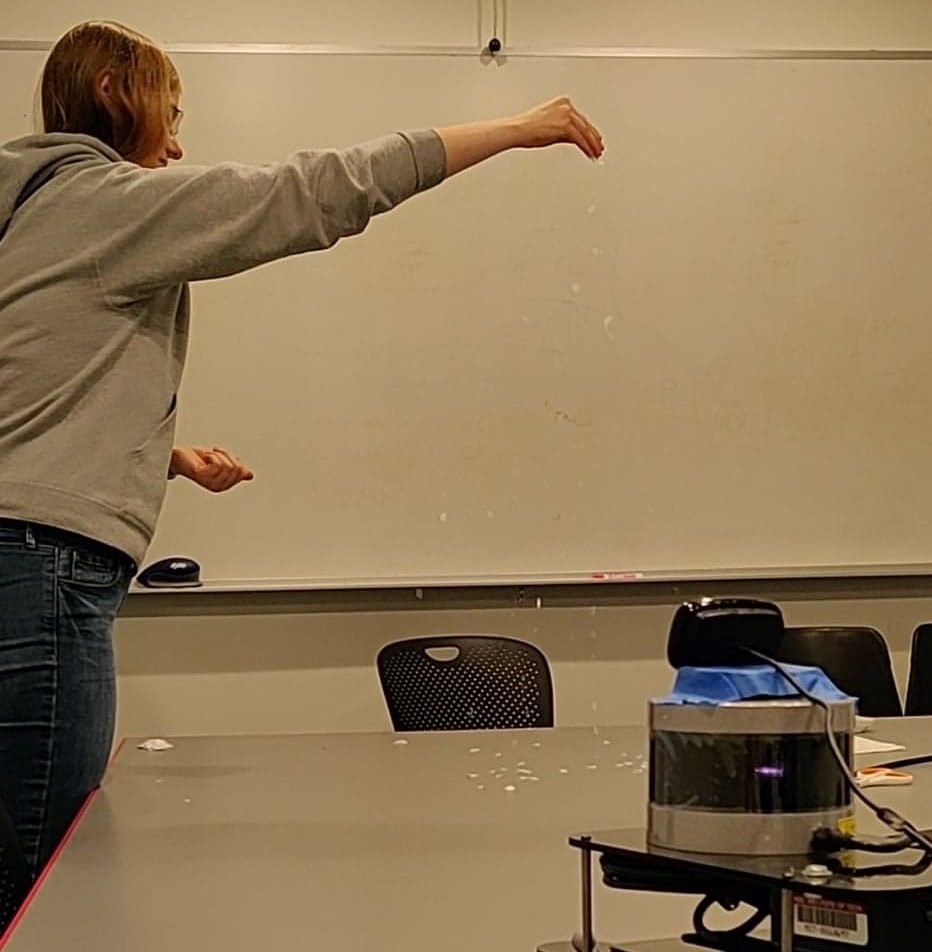}
    \caption{}
  \end{subfigure}
  \begin{subfigure}[t]{0.24\linewidth}
    \includegraphics[width=\textwidth]{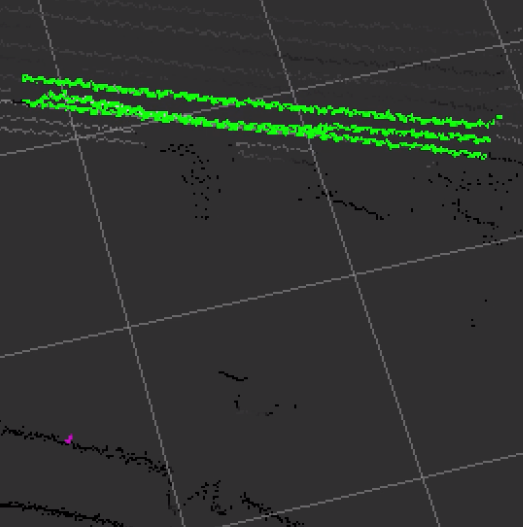}
    \caption{}
  \end{subfigure}
  \begin{subfigure}[t]{0.24\linewidth}
    \includegraphics[width=\textwidth]{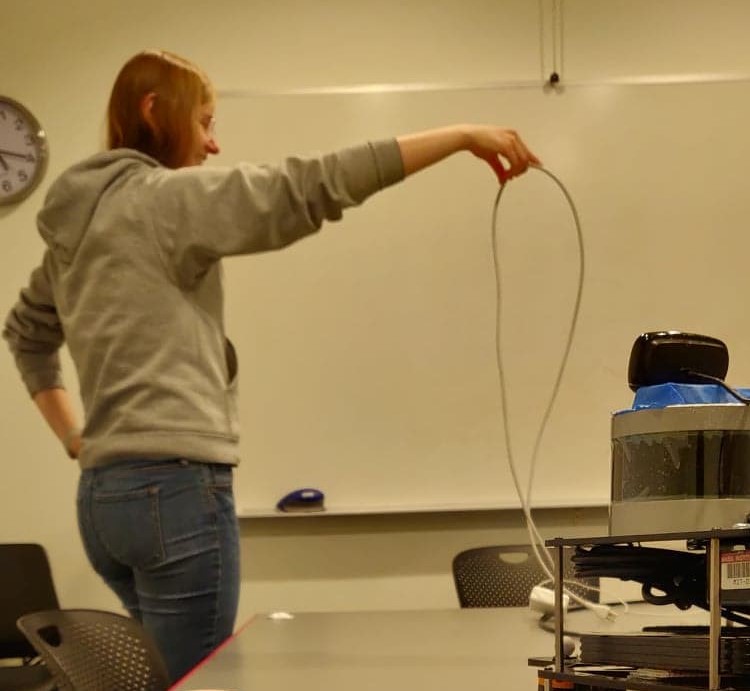}
    \caption{}
  \end{subfigure}
  \begin{subfigure}[t]{0.24\linewidth}
    \includegraphics[width=\textwidth]{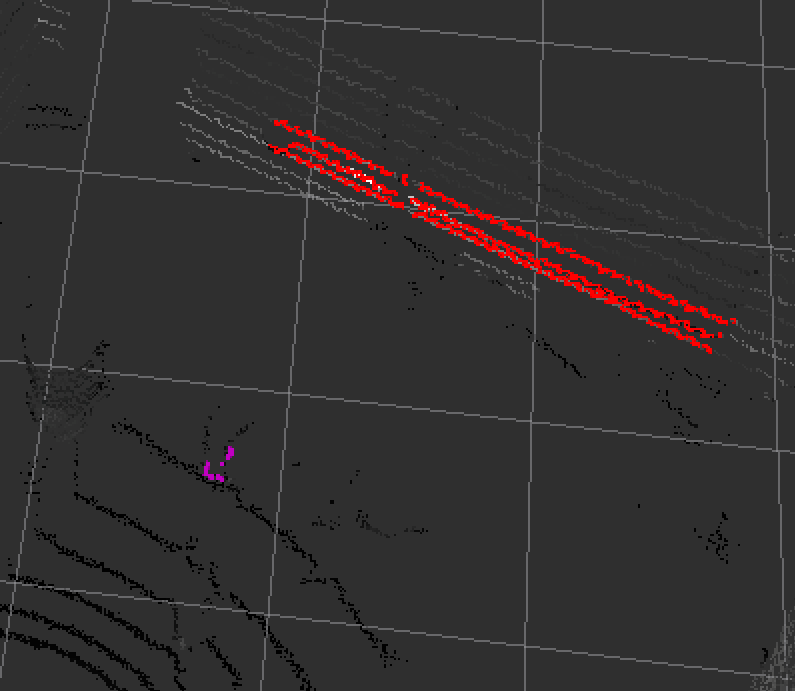}
    \caption{}
  \end{subfigure}
  \caption{(a) Simulating snow with confetti. (b) Certificate points (green) with excluded snow points (purple). (c) Simulating thin obstacles. (d) Certificate points (red) with excluded obstacle points (purple).}
  \label{fig:lidar_test}
\end{figure}

To address \textbf{RQ1}, we tested our monitor's performance in simulated snowy conditions by dropping confetti-like paper in front of the robot car and confirming that the monitor accepts a certificate when there is sufficient space between the car and an obstacle ahead, despite the presence of simulated snow (Fig. \ref{fig:lidar_test}). 



Indeed, when the controller's filtering parameters are set appropriately\footnote{See Section~\ref{sec-benefits}}, it properly identifies points as snow and passes a certificate excluding them to the monitor. Assuming snow is distributed fairly evenly across the lane ahead (and there is not a total ``white out'') the remaining points are still sufficiently dense to establish absence of an obstacle in the lane ahead. 

This establishes some sensitivity of the system. We also tested the converse case, to establish reasonable tolerance. To demonstrate that the monitor would detect obstacles even when the controller fails to do so, we modified the controller so that it would erroneously filter out some sparse but significant obstacles, such as fallen tree branches (simulated, for our robot car, with plastic cables). As expected, since the controller omits points on the obstacle from its certificate, the monitor's horizontal density check fails, the certificate is rejected, and the unsafe action is prevented (Fig \ref{fig:lidar_test}).




To answer \textbf{RQ2}, we measured the size of the various modules of the codebase, comparing in particular the size of the monitor itself (the crucial component of the trusted base) with the size of the controller functions that are used to generate the certificate (and which are \textit{not} part of the trusted base).

Our results (Fig.~\ref{fig:codesizes}) show that, even with a controller performing only basic snow filtering---with no machine learning algorithms---the controller implementation required almost 70 times as many lines of non-comment code as the monitor itself (Fig.~\ref{fig:codesizes}).

\begin{figure}
  \begin{center}
    \caption[]{Comparison of code sizes for LiDAR monitor vs. certificate generation within the controller\footnotemark}
    \begin{tabular}{ | l | c | r |} \hline \rowcolor{Gray}
      Component & Lines of Code \\
      \hline
        
      \bf{LiDAR Monitor Total} & \bf{32}  \\ \hline
      K-d tree library & 739 \\
      Numpy library & 1325 \\
      Certificate generation & 98 \\
      \bf{LiDAR Controller Total}  & \bf{2162} \\
      
      \hline
    \end{tabular}
    \label{fig:codesizes}
  \end{center}
\end{figure}
\footnotetext{These counts were done on a slight variant of the monitor code used for the experimentation (the same variant as is verified in Appendix~\ref{proof-appendix}).}

Finally, to address \textbf{RQ3}, we formally verified the monitor code, constructing a Hoare-style proof by hand (Appendix~\ref{proof-appendix}). Since the code follows the specification very closely, the only (minor) complication is the need for loop invariants.

Although the main complexity gap is due to the snow filtering, more subtle aspects of how the certificate is defined have an impact too. In our initial implementation\footnote{The racecar experiments were performed with this version of the code; the second version is functionally indistinguishable since it just moves the boundary between the controller and the monitor.}, the points in the certificate were not sorted, and the row heights array was not given. This meant that the monitor had to sort the points, and keep track of the maximum and minimum heights of points within each row. Neither involves much code complexity, but by tightening the specification of the certificate so that the monitor can refrain from these additional computations, its complexity was reduced even further and verification became nearly trivial. 

Note that also, importantly, by removing sorting from the monitor, we remove from the trusted base Python's built-in sorting function, which due to its generality and efficiency is non-trivial (and is \textit{not} counted in the lines of code in the table, so the benefit of certified control is actually better than the table suggests).

\subsection{Limitations}

The primary limitation of the LiDAR certificate is that it provides information only about the shape and size of obstacles\footnote{LiDAR sensors also provide information about reflectivity; however this information is not leveraged in our system and is not thought to be critical to obstacle detection.}, yet objects with the same shape and size may present varying degrees of danger. A paper bag, for example, could occlude the same region of points in a LiDAR certificate as a falling rock. Our monitor cannot distinguish between certificates with these same-sized holes.

Another problem is that the certificate only establishes absence of obstacles that have \textit{both} a minimum width and height. A valid certificate could be constructed in the presence of a tall but very skinny object, for example or an inclined one.

These limitations are not limitations of certified control itself, however. If the designer were to devise a scheme for detecting a wider range of obstacles, an argument for the scheme's soundness would presumably be translatable into a certificate. If not, this would cast some doubt on the scheme itself because it would mean that its judgments cannot be straightforwardly justified by concrete evidence.

Another limitation arises from the non-uniformity of the LiDAR data itself. Our LiDAR unit performs its horizontal scans at fixed vertical positions. This means that it is not possible to reduce the vertical spacing to discriminate obstacles at a greater distance from the car. Consequently, thin horizontal obstacles are more problematic than thin vertical obstacles. Again, this is not a problem with certified control itself.




It is important to note that none of these limitations affect the \textit{credibility} of the safety property. Rather, they weaken the safety property itself. Certified control still guarantees that the properties checked by the monitor hold; the limitations involve the implications of those properties for the safety requirement in the physical environment.


\section{Related Work}\label{sec:related}

The literature on safety assurance of vehicle dynamics splits
roughly into two camps, both of which focus on assurance of the planning/control system, and assume perception is assured by some other mechanism.

One focuses on numerical analysis of
reachable states. For example, reachable set computations can justify conflict resolution algorithms that safely allocate disjoint road areas to  traffic participants ~\cite{manzingerTacticalDecisionMaking2018}.
The other focuses on deductive proofs of safety.
The work of~\cite{loosAdaptiveCruiseControl2011} models cars with double integrator
dynamics, and uses the theorem prover KeYmaera to prove safety of a
highway scenario, including lane changing, with arbitrarily many lanes
and arbitrarily many vehicles. The work of~\cite{arechigaUsingTheoremProvers2012,arechigaUsingVerifiedControl2014}
demonstrates how these safety constraints can be used for verification
and synthesis of control policies, including control policies with
switching. In~\cite{loosSafeIntersectionsCrossing2011}, the authors
develop safety contracts that include intersections, and provide
KeYmaera proofs to demonstrate safety.

Responsibility-Sensitive Safety (RSS)~\cite{shalev-shwartzFormalModelSafe2018}
is a framework that assigns
responsibility for safety maneuvers, and ensures that if every traffic
participant meets its responsibilities, no accident will occur. The monitoring of these conditions is conducted as the last phase of planning, and is not separated out as a trusted base. The RSS safety criteria have been explored more formally to boost confidence in their validity~\cite{koopmanAutonomousVehiclesMeet2019}.

All of these works focus on control, and assume that the perception system is reliable. In contrast, certified control takes the perception subsystem out of the trusted base. Nevertheless, these approaches are synergistic with ours. In our design, the low-level controller is still within the trusted base and would benefit from verification. RSS provides more sophisticated runtime criteria than those we have considered that could be incorporated into certificates (e.g., for avoiding the risk of collisions with traffic crossing at an intersection).

Several approaches aim, like ours, to establish safety using some kind of monitor. The Simplex Architecture~\cite{luishaUsingSimplicityControl2001,phanComponentBasedSimplexArchitecture2017}, described above, uses two controllers: a verified safety controller and a performance controller. When safety-critical
situations are detected, the system switches to the verified controller,
but otherwise operates under the performance controller. \cite{phanNeuralSimplexArchitecture2020} extends
Simplex to neural network-based controllers. As we noted, this approach
does not address flaws in perception. In contrast, reasonableness
monitors~\cite{gilpinMonitoringSceneUnderstanders2018,gilpinReasonablenessMonitors2018}
defend against flawed perception by translating the output of a
perception system into relational properties drawn from an ontology that
can be checked against \textit{reasonableness constraints}. This ensures
that the perception system does not make nonsensical inferences, such as
mailboxes crossing the street, but aims for a less complete safety case
than certified control. Similarly, the work of
\cite{kimInterpretableLearningSelfDriving2017} seeks to explain
perception results by analyzing regions of an image that influence the
perception result. These techniques may be useful to enable the
perception system to present pixel regions to the safety monitor as
evidence of a correct prediction.

Other techniques seek to use the safety specifications to automatically
stress test the implementation~\cite{korenAdaptiveStressTesting2018,qinAutomaticTestingFalsification2020,corsoAdaptiveStressTesting2019}. A different approach checks runtime scenarios dynamically against previously executed test suites, generating warnings when the car strays beyond the envelope implicitly defined by those tests~\cite{mauritzAssuringSafetyAdvanced2016}. Certified control is similar in that the certificate criteria represent the operational envelope considered by the designers, and outside that envelope, it will likely not be possible to generate a valid certificate, leading to a safety intervention.

Assurance of perception systems needs to grapple with two key challenges. The first is the difficulty of determining appropriate specifications for perception systems, and the second is with developing scalable reasoning
techniques to ensure that the implementation satisfies its specifications. Reasonableness monitors provide a partial answer to the first. The second is an active area of research: \cite{katzReluplexEfficientSMT2017},
\cite{ehlersFormalVerificationPieceWise2017}, and~\cite{katzMarabouFrameworkVerification2019}, for example, develop efficient
techniques to prove that a deep neural network satisfies a logical specification. While these technologies are promising, their applicability to industrial-scale applications has not yet been demonstrated. Perception is a particularly thorny problem, since it is not clear what properties of a perception system one would want to formally verify.


\section{Conclusions}

To conclude, we highlight some of the key benefits of the approach, explain why some concerns are not flaws, and discuss some limitations, fundamental and minor.

\subsection{Benefits and Non-Concerns}\label{sec-benefits}

The key benefit of certified control is that safety checks that can be performed in conventional architectures but which, until now, have not been regarded as verifiable, can potentially now be verified.

The essence of the scheme is \textit{not} a change in functionality; indeed, the kinds of checks embodied by our monitors are not especially novel. What is important is the \textit{organization} of the functionality, and the concomitant reduction of the trusted base. 


Some readers may be concerned that our safety guarantees are not absolute. The LiDAR certificate, for example, relies on certain arbitrary thresholds that impact the size of obstacles that will be detected. We do not regard this as a flaw, because such tradeoffs are inevitable in setting safety bounds for a complex system. Indeed, the merit of certified control is that it can accommodate whatever thresholds the designer deems appropriate (which might eventually include probabilistic judgments).

One significant difference between a certified control monitor and a conventional monitor is that the former may trigger a safety intervention when the proposed action is in fact safe, because of a bug in the controller that causes it to fail to generate a good certificate. On the one hand, it seems reasonable to prevent the car from proceeding when its code has failed (and indeed, this is routine practice when runtime assertions fail in critical systems\footnote{The immediate cause of the Ariane-5 disaster was an operand error that caused the inertial guidance system to shutdown; whether or not the flight would have succeeded had this error been ignored is not certain.}.
On the other hand, we must admit that it becomes more problematic when safety interventions are themselves risky.

\subsection{Essential Limitations}

Certified control has some essential limitations, which we now analyze. 

\textbf{Justifiable design}. Our most fundamental assumption is that the design of the main controller can be justified by a logical safety case. Prior to the recent advances in machine learning, this would have been uncontroversial: after all, it seems reasonable to insist that any designer should be able to explain why their system works. In contrast, end-to-end machine learning \cite{endtoendml} connects sensors to actuators through a single network. While exciting, the dangers of this approach are clear, since behavior may be unpredictable in scenarios not covered by training data (or in the presence of adversarial attacks \cite{papernotPracticalBlackBoxAttacks2017}).

\textbf{Checking end-to-end safety.} The checking of the certificate is intended to establish an end-to-end safety case; for example, a subset of LiDAR points is taken as evidence of absence of an obstacle allowing the car to continue ahead. This contrasts with a sanity check approach that relies on a more ad hoc collection of partial (and even internal) checks. The problem with our approach is that it is more burdensome to implement; its advantage is that it gives greater confidence, since it is possible for a failing controller to pass a collection of sanity checks. Of course, a wise designer would include sanity checks (for example as runtime assertions) in any case.

\textbf{Dynamic checking.} We assume that it is possible to check safety at runtime by executing a predicate on a certificate comprising sensor data and its interpretation. This may not be possible if the main controller uses algorithms whose correctness on an execution cannot be effectively checked, because the gap between finding and checking is small. For example, when an algorithm that decides whether two graphs are isomorphic returns ``no'' it seems that checking the answer would require enumerating all candidate mappings between the graphs. But even in such cases Blum has shown that probabilistic checks can provide arbitrarily high degrees of confidence \cite{blumDesigningProgramsThat1995}.

\subsection{Non-essential Limitations}

Less essential aspects that suggest further work are:

\textbf{Time independence}. In the current design, each certificate is evaluated independently. In practice, the controller would present a certificate in every controller cycle, and the checking of the certificates would accumulate evidence across frames. This would allow lane lines still to be inferred even though they are periodically obscured by other vehicles.


\textbf{Determinism}. A related limitation is that certificates are currently interpreted deterministically. In practice, it would obviously be desirable to introduce a probabilistic component. For example, a LiDAR certificate might allow areas of insufficient density so long as they do not occur systematically in subsequent frames.

\textbf{Single sensor}. Currently, certificates contain readings from the sensors used by the main controller. An alternative scheme would allow the monitor to obtain sensor readings from an additional sensor of its own. This need not compromise the essential idea that the monitor does not include the interpretation algorithms. For example, a LiDAR-based certificate, instead of including actual signed LiDAR points, might include LiDAR coordinates that tell the monitor where to look to obtain the relevant points.

\bibliographystyle{splncs04}
\bibliography{interlock_paper}

\begin{thebibliography}{10}
\providecommand{\url}[1]{\texttt{#1}}
\providecommand{\urlprefix}{URL }
\providecommand{\doi}[1]{https://doi.org/#1}

\bibitem{TeslaAutopilotDrives}
Tesla {{Autopilot Drives Straight Towards Concrete Barrier}} on {{Highway}}

\bibitem{TestingAutonomousVehicles}
Testing of {{Autonomous Vehicles}}.
  https://www.dmv.ca.gov/portal/dmv/detail/vr/autonomous/testing

\bibitem{TheyWriteRight}
They {{Write}} the {{Right Stuff}}.
  https://www.fastcompany.com/28121/they-write-right-stuff

\bibitem{SafetyManagementRequirements2007}
Safety {{Management Requirements}} for {{Defence Systems}}: {{Part}} 2:
  {{Guidance}} on {{Establishing}} a {{Means}} of {{Complying}} with {{Part
  I}}". Tech. rep., {UK Ministry of Defense} (2007)

\bibitem{CommaAIDriving2020}
Comma {{AI Driving Dataset}}. https://github.com/commaai/comma2k19 (Apr 2020)

\bibitem{CommaaiOpenpilot2020}
Comma {{AI OpenPilot Software}}. https://github.com/commaai/openpilot (Apr
  2020)

\bibitem{arechigaUsingTheoremProvers2012}
Arechiga, N., Loos, S.M., Platzer, A., Krogh, B.H.: Using theorem provers to
  guarantee closed-loop system properties. In: 2012 {{American Control
  Conference}} ({{ACC}}). pp. 3573--3580. {IEEE}, {Montreal, QC} (Jun 2012)

\bibitem{arechigaUsingVerifiedControl2014}
Arechiga, N., Krogh, B.H.: Using verified control envelopes for safe controller
  design. In: American {{Control Conference}} (2014)

\bibitem{verified-tls}
Bhargavan, K., Fournet, C., Corin, R., Z\u{a}linescu, E.: Verified
  cryptographic implementations for tls. ACM Trans. Inf. Syst. Secur.
  \textbf{15}(1) (Mar 2012). \doi{10.1145/2133375.2133378},
  \url{https://doi.org/10.1145/2133375.2133378}

\bibitem{blumDesigningProgramsThat1995}
Blum, M., Kannan, S.: Designing {{Programs}} that {{Check Their Work}}. Journal
  of the ACM  (1995)

\bibitem{endtoendml}
Bojarski, M., Del~Testa, D., Dworakowski, D., Firner, B., Flepp, B., Goyal, P.,
  Jackel, L.D., Monfort, M., Muller, U., Zhang, J., Zhang, X., Zhao, J., Zieba,
  K.: End to {{End Learning}} for {{Self}}-{{Driving Cars}}. arXiv:1604.07316
  [cs]  (Apr 2016)

\bibitem{cameronDriverlessReadinessScore}
Cameron, O.: The {{Driverless Readiness Score}}.
  https://olivercameron.substack.com/p/the-driverless-readiness-score, library
  Catalog: olivercameron.substack.com

\bibitem{charronDenoisingLidarPoint2018}
Charron, N., Phillips, S., Waslander, S.L.: De-noising of {{Lidar Point Clouds
  Corrupted}} by {{Snowfall}}. In: Computer and {{Robotic Vision}} (2018)

\bibitem{corsoAdaptiveStressTesting2019}
Corso, A., Du, P., {Driggs-Campbell}, K., Kochenderfer, M.J.: Adaptive {{Stress
  Testing}} with {{Reward Augmentation}} for {{Autonomous Vehicle Validation}}.
  In: {{IEEE Intelligent Transportation Systems Conference}} (2019)

\bibitem{crenshawSimplexReferenceModel2007}
Crenshaw, T.L., Gunter, E., Robinson, C.L., Sha, L., Kumar, P.R.: The simplex
  reference model: {{Limiting}} fault- propagation due to unreliable components
  in cyber-physical system architectures. In: {{IEEE International
  Real}}-{{Time Systems Symposium}} (2007)

\bibitem{10.5555/1792734.1792766}
De~Moura, L., Bj\o{}rner, N.: Z3: An efficient smt solver. In: Proceedings of
  the Theory and Practice of Software, 14th International Conference on Tools
  and Algorithms for the Construction and Analysis of Systems. p. 337–340.
  TACAS’08/ETAPS’08, Springer-Verlag, Berlin, Heidelberg (2008)

\bibitem{ehlersFormalVerificationPieceWise2017}
Ehlers, R.: Formal {{Verification}} of {{Piece}}-{{Wise Linear Feed}}-{{Forward
  Neural Networks}}. arXiv:1705.01320 [cs]  (Aug 2017)

\bibitem{fischlerRandomSampleConsensus1981}
Fischler, M.A., Bolles, R.C.: Random sample consensus: A paradigm for model
  fitting with applications to image analysis and automated cartography.
  Communications of the ACM  (1981)

\bibitem{gilpinReasonablenessMonitors2018}
Gilpin, L.H.: Reasonableness {{Monitors}}. AAAI  (2018)

\bibitem{gilpinMonitoringSceneUnderstanders2018}
Gilpin, L.H., Macbeth, J.C.: Monitoring {{Scene Understanders}} with
  {{Conceptual Primitive Decomposition}} and {{Commonsense Knowledge}}.
  Advances in Cognitive Systems p.~20 (2018)

\bibitem{greenleeDriverVigilanceAutomated2018}
Greenlee, E.T., DeLucia, P., Newton, D.C.: Driver {{Vigilance}} in {{Automated
  Vehicles}}: {{Hazard Detection Failures Are}} a {{Matter}} of {{Time}}. Human
  Factors  (2018)

\bibitem{jacksonSoftwareDependableSystems2007}
Jackson, D., Thomas, M., Millett, L.I. (eds.): Software for {{Dependable
  Systems}}: {{Sufficient Evidence}}? {National Research Council} (2007)

\bibitem{kalraDrivingSafetyHow2016}
Kalra, N., Paddock, S.M.: Driving to {{Safety}}: {{How Many Miles}} of
  {{Driving Would It Take}} to {{Demonstrate Autonomous Vehicle Reliability}}?
  Tech. Rep. RAND RR-1478-RC, {RAND Corporation} (2016)

\bibitem{katzMarabouFrameworkVerification2019}
Katz, G., et. al.: The {{Marabou Framework}} for {{Verification}} and
  {{Analysis}} of {{Deep Neural Networks}}. In: Dillig, I., Tasiran, S. (eds.)
  Computer {{Aided Verification}}, vol. 11561, pp. 443--452. {Springer
  International Publishing}, {Cham} (2019), series Title: Lecture Notes in
  Computer Science

\bibitem{katzReluplexEfficientSMT2017}
Katz, G., Barrett, C., Dill, D., Julian, K., Kochenderfer, M.: Reluplex: {{An
  Efficient SMT Solver}} for {{Verifying Deep Neural Networks}}.
  arXiv:1702.01135 [cs]  (May 2017)

\bibitem{kimInterpretableLearningSelfDriving2017}
Kim, J., Canny, J.: Interpretable {{Learning}} for {{Self}}-{{Driving Cars}} by
  {{Visualizing Causal Attention}}. In: 2017 {{IEEE International Conference}}
  on {{Computer Vision}} ({{ICCV}}). pp. 2961--2969. {IEEE}, {Venice} (Oct
  2017)

\bibitem{kleinSeL4FormalVerification2009}
Klein, G., et. al.: {{seL4}}: Formal verification of an {{OS}} kernel. In:
  Proceedings of the {{ACM SIGOPS}} 22nd Symposium on {{Operating}} Systems
  Principles - {{SOSP}} '09. p.~207. {ACM Press}, {Big Sky, Montana, USA}
  (2009)

\bibitem{koopmanAutonomousVehiclesMeet2019}
Koopman, P., Osyk, B., Weast, J.: Autonomous {{Vehicles Meet}} the {{Physical
  World}}: {{RSS}}, {{Variability}}, {{Uncertainty}}, and {{Proving Safety}}.
  In: Romanovsky, A., Troubitsyna, E., Bitsch, F. (eds.) Computer {{Safety}},
  {{Reliability}}, and {{Security}}, vol. 11698, pp. 245--253. {Springer
  International Publishing}, {Cham} (2019), series Title: Lecture Notes in
  Computer Science

\bibitem{korenAdaptiveStressTesting2018}
Koren, M., Alsaif, S., Lee, R., Kochenderfer, M.J.: Adaptive stress testing for
  autonomous vehicles. In: {{IEEE Intelligent Vehicles Symposium}} (2018)

\bibitem{loosSafeIntersectionsCrossing2011}
Loos, S.M., Platzer, A.: Safe intersections: {{At}} the crossing of hybrid
  systems and verification. In: 2011 14th {{International IEEE Conference}} on
  {{Intelligent Transportation Systems}} ({{ITSC}}). pp. 1181--1186. {IEEE},
  {Washington, DC, USA} (Oct 2011)

\bibitem{loosAdaptiveCruiseControl2011}
Loos, S.M., Platzer, A., Nistor, L.: Adaptive {{Cruise Control}}: {{Hybrid}},
  {{Distributed}}, and {{Now Formally Verified}}. In: Butler, M., Schulte, W.
  (eds.) {{FM}} 2011: {{Formal Methods}}, vol.~6664, pp. 42--56. {Springer
  Berlin Heidelberg}, {Berlin, Heidelberg} (2011), series Title: Lecture Notes
  in Computer Science

\bibitem{luishaUsingSimplicityControl2001}
{Lui Sha}: Using simplicity to control complexity. IEEE Software
  \textbf{18}(4),  20--28 (Jul 2001)

\bibitem{manzingerTacticalDecisionMaking2018}
Manzinger, S., Althoff, M.: Tactical {{Decision Making}} for {{Cooperative
  Vehicles Using Reachable Sets}}. In: 2018 21st {{International Conference}}
  on {{Intelligent Transportation Systems}} ({{ITSC}}). pp. 444--451. {IEEE},
  {Maui, HI} (Nov 2018)

\bibitem{mauritzAssuringSafetyAdvanced2016}
Mauritz, M., Howar, F., Rausch, A.: Assuring the {{Safety}} of {{Advanced
  Driver Assistance Systems Through}} a {{Combination}} of {{Simulation}} and
  {{Runtime Monitoring}}. In: Margaria, T., Steffen, B. (eds.) Leveraging
  {{Applications}} of {{Formal Methods}}, {{Verification}} and {{Validation}}:
  {{Discussion}}, {{Dissemination}}, {{Applications}}, vol.~9953, pp. 672--687.
  {Springer International Publishing}, {Cham} (2016), series Title: Lecture
  Notes in Computer Science

\bibitem{nairRectifiedLinearUnits2010}
Nair, V., Hinton, G.E.: Rectified {{Linear Units Improve Restricted Boltzmann
  Machines}}. In: International {{Conference}} on {{Machine Learning}}. p.~8
  (2010)

\bibitem{papernotPracticalBlackBoxAttacks2017}
Papernot, N., McDaniel, P., Goodfellow, I., Jha, S., Celik, Z.B., Swami, A.:
  Practical {{Black}}-{{Box Attacks}} against {{Machine Learning}}. Asia
  Conference on Computer Science and Security  (Mar 2017)

\bibitem{phanComponentBasedSimplexArchitecture2017}
Phan, D., et. al.: A {{Component}}-{{Based Simplex Architecture}} for
  {{High}}-{{Assurance Cyber}}-{{Physical Systems}}. 2017 17th International
  Conference on Application of Concurrency to System Design (ACSD) pp. 49--58
  (Jun 2017)

\bibitem{phanNeuralSimplexArchitecture2020}
Phan, D.T., Grosu, R., Jansen, N., Paoletti, N., Smolka, S.A., Stoller, S.D.:
  Neural {{Simplex Architecture}}. arXiv:1908.00528 [cs, eess]  (Mar 2020)

\bibitem{pulinaChallengingSMTSolvers2012}
Pulina, L., Tacchella, A.: Challenging {{SMT}} solvers to verify neural
  networks. AI Communications  \textbf{25}(2),  117--135 (2012)

\bibitem{qinAutomaticTestingFalsification2020}
Qin, X., Ar{\'e}chiga, N., Best, A., Deshmukh, J.: Automatic {{Testing}} and
  {{Falsification}} with {{Dynamically Constrained Reinforcement Learning}}.
  arXiv:1910.13645 [cs, eess]  (Feb 2020)

\bibitem{saltzerEndtoendArgumentsSystem1984}
Saltzer, J.H., Reed, D.P., Clark, D.D.: End-to-end arguments in system design.
  ACM Transactions on Computer Systems (TOCS)  \textbf{2}(4),  277--288 (Nov
  1984)

\bibitem{shalev-shwartzFormalModelSafe2018}
{Shalev-Shwartz}, S., Shammah, S., Shashua, A.: On a {{Formal Model}} of
  {{Safe}} and {{Scalable Self}}-driving {{Cars}}. arXiv:1708.06374 [cs, stat]
  (Oct 2018)

\bibitem{steinMitigationStrategiesDesign2007}
Stein, W.J., Neuman, T.R.: Mitigation strategies for design exceptions. Tech.
  rep., {United States. Federal Highway Administration. Office of Safety}
  (2007)

\bibitem{wakabayashiSelfDrivingUberCar2018}
Wakabayashi, D.: Self-{{Driving Uber Car Kills Pedestrian}} in {{Arizona}},
  {{Where Robots Roam}}. The New York Times  (Mar 2018)

\bibitem{weinstockDependabilityCases2004}
Weinstock, C.B., Goodenough, J.B., Hudak, J.J.: Dependability {{Cases}}. Tech.
  Rep. CMU/SEI-2004-TN-016, {CMU Software Engineering Institute} (2004)

\bibitem{yehDependability777Primary1995}
Yeh, Y.C.: Dependability of the 777 {{Primary Flight Control System}}. In:
  {{Dependable Computing for Critical Applications}} (1998)

\end{thebibliography}

\newpage
\appendix
\section{Architecture}\label{architecture-appendix}

\begin{figure}
  \begin{center}
  \includegraphics[width=0.95\textwidth]{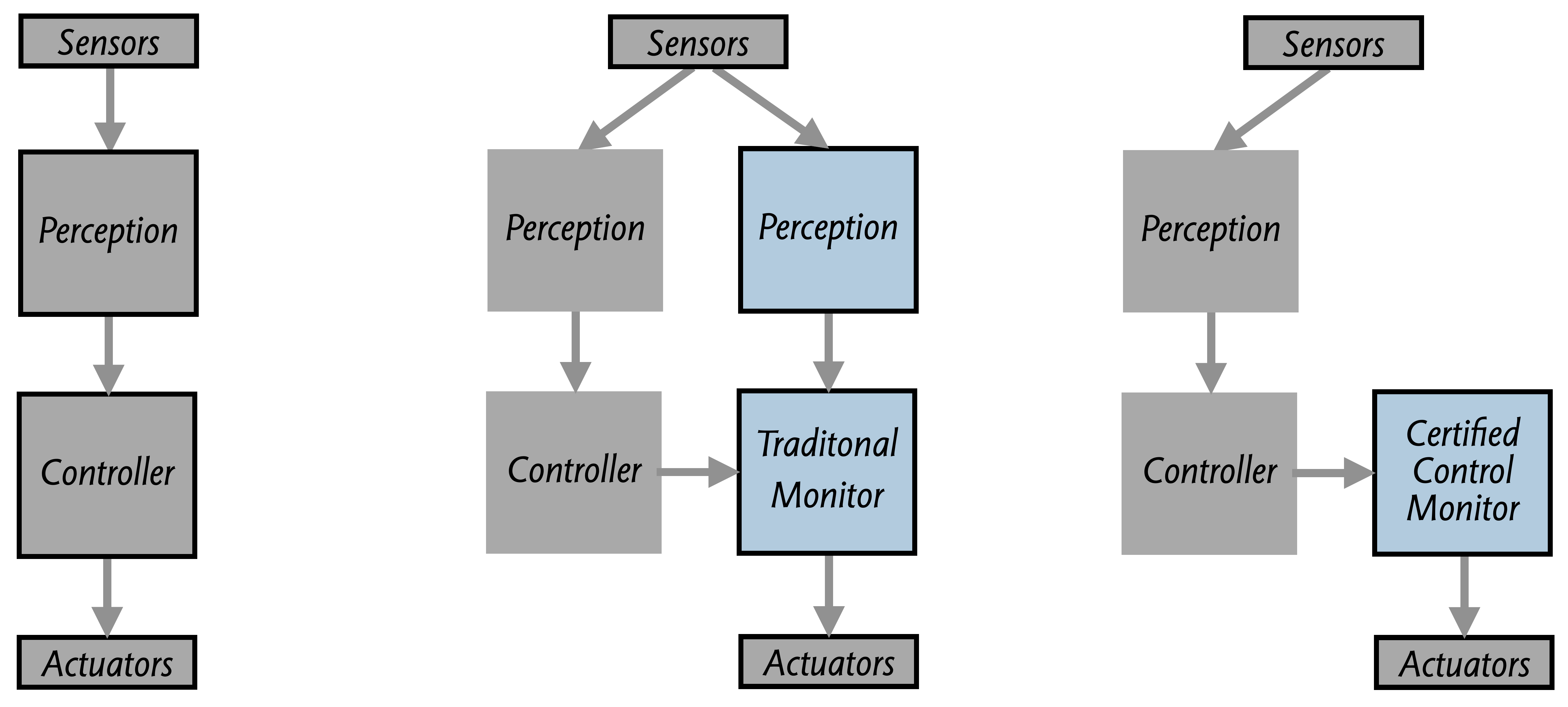}
  \end{center}
  
  \caption{
  Architectures for safe control. On the left, without a monitor. In the center, with a traditional monitor. On the right, with a certified control monitor. Components of the trusted base are shown with bold outlines.}
\label{fig:certified_control_diagrams}
\end{figure}

 Figure~\ref{fig:certified_control_diagrams} highlights the difference between a certified control monitor and a traditional monitor. On the left, it shows an architecture without a monitor, in which a perception module reads and interprets sensor data, and passes the results to a controller, which then decides on an appropriate action that is passed to the actuators. In the center is the architecture for a traditional monitor that performs its own perception. On the right is the architecture with a certified control monitor, which delegates to the controller the task of assembling perception results into compelling evidence in the certificate.
 
\section{LiDAR Certificate Details}\label{proof-appendix}

In this section, we give the details of the LiDAR certificate and its checking. Figure~\ref{fig:params} shows the parameters comprising the certificate.
Figure~\ref{fig:spec} gives the specification of the certificate: that is, the validity rules checked by the monitor.


\begin{figure}
\begin{Verbatim}[fontsize=\scriptsize,frame=single]
A LiDAR point is a tuple (fb, rl, ud), with each coordinate in meters,
with (0, 0, 0) centered on the LiDAR scanner (forward/backward, right/left,
up/down: positive in direction of first of each). All distances in metres,
measured from centerpoint

min_forward_dist: minimum acceptable distance for all points
lane_left, lane_right: left and right lane edges at distance min_forward_dist
lane_up, lane_down: required spread of points vertically
max_rl_diff: maximum horizontal spacing of points within each row
max_ud_diff: maximum vertical spacing of rows
max_row_dev: maximum vertical deviation of point in row from row height
row_heights: array of vertical heights for rows
rows: array of rows, each being an array of points
\end{Verbatim}
  \caption{
  Parameters comprising the LiDAR certificate and their informal definitions}
\label{fig:params}
\end{figure}

\begin{figure}
\begin{Verbatim}[fontsize=\scriptsize,frame=single]
Result is conjunction of:
Min forward distance conformance
  all i: 0..len(rows)-1 | all j: 0..len(rows[i])-1 | FDC(i,j)
  where FDC(i,j) = fb(rows[i][j]) >= min_forward_dist
Row height conformance
  all i: 0..len(rows)-1 | all j: 0..len(rows[i])-1 | RHC(i,j)
  where RHC(i,j) = abs (ud(proj(rows[i][j])) - row_heights[i]) <= max_row_dev
Row height separation
  all i: 0..len(rows)-2 | RHS(i)
  where RHS(i) = abs(row_heights[i] - row_heights[i+1]) <= max_ud_diff
Horizontal density
  all i: 0..len(rows)-1 | all j: 0..len(rows[i])-2 | HD(i,j)
  where HD(i,j) = abs (rl(proj(rows[i][j])) - rl(proj(rows[i][j+1])) <= max_rl_diff
Horizontal spread
  all i: 0..len(rows)-1 | HS(i)
  where HS(i) = rl(proj(rows[i][0])) <= lane_left and
        rl(proj(rows[i][len(rows[i])-1])) >= lane_right
Vertical spread (assumes row heights ordered from top to bottom)
  LU where LU = row_heights[0] >= lane_up
  LD where LD = row_heights[len(row_heights)-1] <= lane_down
where proj(p) =
  let k = min_forward_dist / fb(p) in (min_dist, rl(p) * k, ud(p) * k)
  in (min_forward_dist, rl(p) * k, ud(p) * k)
and ud(p) = p[2], rl(p) = p[1] and fb(p) = p[0]
\end{Verbatim}
\caption{
  Specification of the LiDAR certificate}
\label{fig:spec}
\end{figure}

\clearpage

\subsection{Code Meets Specification}

Python code for the monitor itself follows, which includes Hoare-style annotations that comprise a proof of correctness. We used Python because it is the preferred language for both our simulator and race car environment. In a real installation of certified control, we would expect a more appropriate language to be used instead, for example, a safe subset of C (such as MISRA-C) or a real-time language designed for verification (such as a SPARK Ada). Using Python would impose additional proof obligations---for example, that the certificate parameters have the appropriate types---which we did not include here.

The assertions are sufficiently complete that a Hoare-style proof follows directly from them, requiring only trivial verification conditions (eg, that the assertion immediately prior to each loop implies its invariant). The assertions establish only partial correctness. Termination is proved with the decrementing function {\tt len(a)-i} for each loop in which index {\tt i} is used to iterate over array {\tt a}.

\begin{Verbatim}[fontsize=\scriptsize,
frame=single,
numbers=left,
label=Python code for the monitor that checks the certificate\, annotated with invariants]
def interlock (min_forward_dist, lane_left, lane_right, lane_up, lane_down,
max_rl_diff, max_ud_diff, max_row_dev, row_heights, rows):

# define functions to extract up/down, right/left and forward/backward
# coordinates from points
def ud (p):
  return p[2]
def rl (p):
  return p[1]
def fb (p):
  return p[0]

# return a point corresponding to projection onto plane at min_forward_dist
def proj(p):  k = min_forward_dist / fb(p)
  return (min_forward_dist, rl(p) * k, ud(p) * k)

result = True

# vertical spread
result = result && row_heights[0] >= lane_up
result = result && row_heights[len(row_heights)-1] <= lane_down
{result = LU and LD}

i = 0
{i = 0 and result = LU and LD}
while (i < len(rows)):
  {
  i <= len(rows)
  result = LU and LD and
  and all I: 0..i-1 | all J:  0..len(rows[I])-1 | FDC(I,J) and RHC(I,J)
  and all I: 0..i-1 | all J:  0..len(rows[I])-2 | HD(I,J)
  and all I: 0..i-1 | HS(I)
  }

  # horizontal spread
  result = result && rl(proj (rows[i][0])) <= lane_left
  result = result && rl(proj (rows[i][-1])) >= lane_right

  {
  result = LU and LD and
  and all I: 0..i-1 | all J:  0..len(rows[I])-1 | FDC(I,J) and RHC(I,J)
  and all I: 0..i-1 | all J:  0..len(rows[I])-2 | HD(I,J)
  and all I: 0..i | HS(I)
  }

  j = 0
  while (j < len(rows[i])):
    {
    j <= len(rows[i])
    result = LU and LD and
    and all I: 0..i-1 | all J:  0..len(rows[I])-1 | FDC(I,J) and RHC(I,J)
    all J:  0..j-1 | FDC(i,J) and RHC(i,J)
    and all I: 0..i-1 | all J:  0..len(rows[I])-2 | HD(I,J)
    and all I: 0..i | HS(I)
    }

    # min forward dist conformance
    result = result && fb(rows[i][j]) >= min_forward_dist

    # row height conformance
    dev = abs( ud(proj(rows[i][j])) - row_heights[i] )
    result = result && dev <= max_row_dev

    j++

  {
  result = LU and LD and
  and all I: 0..i | all J:  0..len(rows[I])-1 | FDC(I,J) and RHC(I,J)
  and all I: 0..i-1 | all J:  0..len(rows[I])-2 | HD(I,J)
  and all I: 0..i | HS(I)
  }

  j = 0
  while (j < len(rows[i])-1)):
    {
    j <= len(rows[i])-1
    result = LU and LD and
    and all I: 0..i | all J:  0..len(rows[I])-1 | FDC(I,J) and RHC(I,J)
    and all I: 0..i-1 | all J:  0..len(rows[I])-2 | HD(I,J)
    and all J:  0..j-1 | HD(i,j)
    and all I: 0..i | HS(I)
    }
     # horizontal density
    rl_diff = abs( rl(proj(rows[i][j])) - rl(proj(rows[i][j+1])))
     result = result && rl_diff <= max_rl_diff
    j++
  i++
{
result = LU and LD and
and all i: 0..len(rows)-1 | all j: 0..len(rows[i])-1 | FDC(i,j) and RHC(I,J)
and all i: 0..len(rows)-1 | all j: 0..len(rows[i])-2 | HD(i,j)
and all i: 0..len(rows)-1 | HS(i)
}

i = 0
{i = 0}
while (i < len(rows)-1):
  {
  result = LU and LD and
  and all i: 0..len(rows)-1 | all j: 0..len(rows[i])-1 | FDC(i,j) and RHC(I,J)
  and all i: 0..len(rows)-1 | all j: 0..len(rows[i])-2 | HD(i,j)
  and all i: 0..len(rows)-1 | HS(i)
  and all I: 0..i-1 | RHS(I)
  }
  # row height separation
  ud_diff = abs(row_heights[i] - row_heights[i+1])
  result = result && ud_diff <= max_ud_diff
  i++

{
result = LU and LD and
and all i: 0..len(rows)-1 | all j: 0..len(rows[i])-1 | FDC(i,j) and RHC(I,J)
and all i: 0..len(rows)-1 | all j: 0..len(rows[i])-2 | HD(i,j)
and all i: 0..len(rows)-1 | HS(i)
and all i: 0..len(rows)-2 | RHS(i)
}
return result
\end{Verbatim}
\subsection{Specification Implies Requirement}

The safety case must include a proof that the certificate specification establishes the desired safety requirement in the real world, which we outline below.

\textit{Requirement}. There is no obstacle closer than min\_forward\_dist
with a horizontal dimension larger than max\_rl\_diff
and a vertical dimension larger than max\_ud\_diff + max\_row\_dev
in the rectangle defined by lane\_left, lane\_right, lane\_up and lane\_down

\textit{Proof that Specification implies Requirement}.
If there were an obstacle violating the requirement closer than \textit{min\_forward\_dist}, it would overlap one of the existing points by the line segment lemma (below), and then the forward/backward component of that LiDAR point would be incorrect.

\textit{Line segment lemma}. Consider a one dimensional line, and a collection of points on it $x[0] \dots x[n-1]$.
Suppose for some $\delta$ and $left < right$ we have
\begin{enumerate}
\item $| x[0] - left | \leq \delta$
\item $| x[n-1] - right | \leq \delta$
\item $\forall i: 0.. n-2 : | x[i] - x[i+1] | \leq \delta$
\end{enumerate}
Then every line segment of $length > \delta$ between $left$ and $right$ includes at least one $x[i]$:

$\forall x, x' : x' - x > \delta \wedge x' < right \wedge x > left \Rightarrow \exists i : x \leq  x[i] \leq x'$

\textit{Proof}.
If a line segment starting at $x$ and ending at $x'$ has $length > \delta$, then $x' > x + \delta$.  If it begins between \textit{left} and $x[0]$ (\textit{left} $\leq x \leq x[0]$),
then $x' > x[0]$ by (1), so $x \leq x[0] \leq x'$. If the line segment ends between $x[n-1]$ and right ($x[n-1] \leq x' \leq
right$), then $x < x[n-1]$ by (2) and $x \leq x[n-1] \leq x'$. If it begins between points  $x[i]$ and $x[i+1]$  ($x[i] \leq x \leq x[i+1]$) then it must end
at a point after $x[i+1]$ by (3), and $x \leq x[i+1] \leq x'$.

Using the Z3 theorem prover~\cite{10.5555/1792734.1792766}, we checked the lemma mechanically, bounding the analysis up to a fixed number of points. In the
following code, the function \texttt{generate\_VC(n)} generates the
verification condition for the case where the number of points is
$n$. At the bottom of the code, it uses Z3 to check if there is a
counterexample violating \texttt{generate\_VC(n)} for all $n \in [3,
maxN)$. Here, we use \texttt{maxN = 100}. But it works for other
constants as well.
\begin{Verbatim}[fontsize=\scriptsize,
frame=single,
numbers=left,
label=Python code for proving the line segment lemma using Z3 theorem prover]
from z3 import *  # Need run `pip install z3-solver`. Tested with z3-solver 4.8.8.0

def generate_VC(n):
  delta = Real('delta')
  left = Real('left')
  right = Real('right')
  assumptions = []
  assumptions.append(delta > 0)
  assumptions.append(left < right)

  x = Reals(' '.join(['x' + str(i) for i in range(n)]))

  # Assumption 1
  assumptions += [-delta <= x[0] - left, x[0] - left <= delta]

  # Assumption 2
  assumptions += [-delta <= x[n - 1] - right, x[n - 1] - right <= delta]

  # Assumption 3
  assumptions += [-delta <= x[i] - x[i+1] for i in range(n - 1)]
  assumptions += [x[i] - x[i+1] <= delta for i in range(n - 1)]
  assumptions = And(assumptions)

  x_L = Real('x_L')
  x_R = Real('x_R')
  antecedent = And(x_R - x_L > delta, x_R < right, x_L > left)
  # Instead of using `Exists`, we enumerate each case for i explicitly.
  consequent = Or([And(x_L <= x[i], x[i] <= x_R) for i in range(n)])

  # We check if there is a model satisfying the *negation* of the lemma.
  return Not(Implies(And(assumptions, antecedent), consequent))

# Check all cases for n in [3, maxN).
maxN = 100
for n in range(3, maxN):
  # It prints out "no solution" indicating the absence of counterexamples.
  solve(generate_VC(n))
\end{Verbatim}

Since Python does not provide type checking, a malicious controller could conceivably create a passable certificate with incorrectly typed data, as long as that data still supports all of the checking operations performed by the monitor. We expect a production system to be implemented in a language with type checking. Further, LiDAR data authentication, described next, prevents this problem.

\subsection{LiDAR Authentication}

Our experimental implementation does not currently include the authentication of LiDAR points required to ensure that the controller does not corrupt or fake them. Since the authentication checks are point-wise, we do not anticipate this being problematic (since it seems to be well within the scope of cryptographic verification \cite{verified-tls}).

We did conduct some timing experiments to ensure that the authentication overhead would not be excessive. Table~\ref{crypto-timing-table} shows timings for a variety of schemes that we tested. As expected, full-blown public-key encryption using RSA or DSS is not viable, although a scheme based on Edwards curves (Ed25519) might be. In practice, a simple hashing scheme with a shared secret is probably the best approach. It allows 100 LiDAR points to be signed and verified in just over a millisecond. Its only disadvantage is that it would require a setup protocol in which the LiDAR sensor unit and the certified control monitor exchange the shared secret.

\begin{table}
  \begin{center}
    \begin{tabular}{ | l | c | c |}
    \hline
Scheme \cellcolor{Gray} & Sign 100 \cellcolor{Gray} & Sign and verify \cellcolor{Gray} \\
 \cellcolor{Gray} & points (ms)  \cellcolor{Gray} &  100 points (ms) \cellcolor{Gray}
\cellcolor{Gray}

                 \\ \hline
      Hashing with secret & 0.6 & 1.4  \\ \hline
      Hashed RSA & 501.8 & 506.1  \\ \hline

      Ed25519 & 5.6 & 18.1 \\ \hline
      DSS & 296.9 & 805.5 \\ \hline
    \end{tabular}
    \caption{Signature scheme runtimes.} \label{crypto-timing-table}
  \end{center}
\end{table}


\section{A certificate for visual lane detection}\label{lane-appendix}

A key part of virtually all autonomous vehicle perception
systems is the camera sensor, since most critical road features such
as lane lines, traffic signs, and traffic signals can only be
identified through visual means. It follows that the software
that identifies these features in the vehicle's camera feed is
enormously complex and difficult to verify by any means other than
exhaustive testing due to the nature of neural nets. Given how difficult it is to test such an important part of the vehicle, it is no surprise that failures in vision perception often lead to fatal crashes.
Thus, being able to provide a safety guarantee for vision perception systems, even for just one feature the system claims to detect, would be an important step towards improving overall safety.

To explore the application of certified control in the domain of
vision, we focused on the verification of lane line detection---that is, we designed a certificate that confirms that the
perception system correctly identifies the location of lane lines at
runtime.

The certificate contains the following elements:
\begin{itemize}
\item the image frame from which the lane lines were deduced;
\item the position of the left and right lane boundaries as
  second-degree polynomials in the bird's-eye/top-down view ($L(i)$
  and $R(i)$, respectively), both lines giving the distance from the
  left side of the top-down view as a function of distance from the front of
  the car;
\item the bird's-eye transformation matrix $T$ used to transform the
  lane lines;
\item a series of color filtering thresholds used to process the image
  and highlight the presence of the lane lines.
\end{itemize}

The monitor verifies the certificate in two checks. The first check is
geometric. In this test, the proposed lane lines are evaluated as a
pair; if they conform to specific geometric bounds (such as
parallelism) then the lane lines pass this test. The second check is a
computer vision check. Computer vision techniques are used to
determine whether the proposed lane lines correspond to lane line
markings in the image. If they do, the lane lines pass. The proposed
lane lines must pass both checks in order for the certificate to be
accepted.

\subsection{Geometric Test}

The geometric test ensures that the purported lane lines are parallel
and spaced according to local regulations; in the US,
for example, the width of a freeway lane is 12 feet~\cite{steinMitigationStrategiesDesign2007}.
To conduct the parallelism test, the monitor checks
that the distance between points on $L(i)$ and the intersection of its
normal with $R(i)$ remains relatively constant. To ensure correct
spacing, this distance is required to be on average within a certain
delta of the expected distance between lane lines.

The problem of checking the lane width and comparing lane boundaries
for parallelism is greatly simplified by working with a bird's-eye
perspective of the lane boundaries. The transformation and fitting of the lane-lines is done by the main controller, so the checks performed
by the monitor remain simple.

\subsection{Conformance Test}

\begin{figure}
  \centering
  \begin{subfigure}{0.49\linewidth}
    \includegraphics[width=\textwidth]{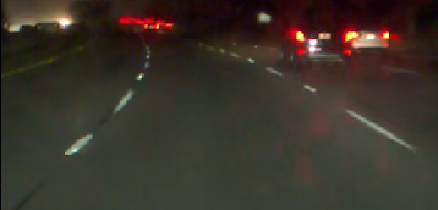}
    \caption{Original image}
  \end{subfigure}
  \begin{subfigure}{0.49\linewidth}
    \includegraphics[width=\textwidth]{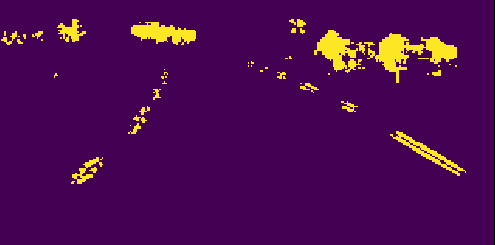}
    \caption{Edge \& lightness filtering}
  \end{subfigure}
  \\
  \begin{subfigure}{0.49\linewidth}
    \includegraphics[width=\textwidth]{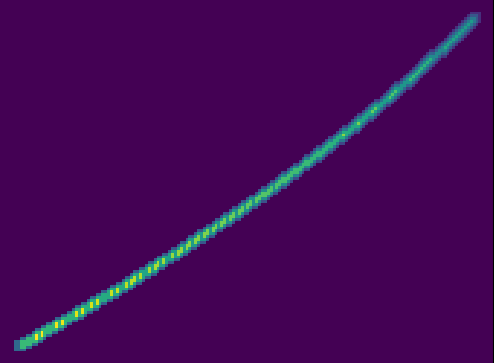}
    \caption{Filter
    %
    for left lane line}
  \end{subfigure}
  \begin{subfigure}{0.49\linewidth}
    \includegraphics[width=\textwidth]{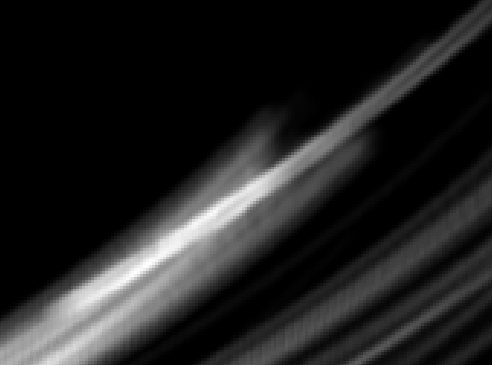}
    \caption{Correlating filter and image}
  \end{subfigure}
  \caption{The steps of the conformance test.}
  \label{fig:conformance_test}
\end{figure}

Proposed lane lines might happen to pass the geometric test (or be
maliciously tailored to pass it) but not actually correspond to lane
lines on the road itself. It is therefore essential to also check that
the lane lines conform to a signed image of the road. To do this, the
monitor applies simple and well-tested computer vision algorithms to
determine if there are corresponding lane markings in the
image. The match between the proposed lane lines and the image is checked as follows (and illustrated in Fig.~\ref{fig:conformance_test})
\footnote{
Our original monitor implementation used a fixed bird's-eye
transformation matrix to convert the camera image to a top-down view
(instead of converting the lane lines from the top-down to camera
view). This allowed us to create an additional filter to specifically
test for dotted lane lines. However, this implementation was less robust to curvy
roads and varying camera angles. In addition, \textit{openpilot} already
computes a calibration matrix for converting lane lines from top-down
view to camera view. We took advantage of this by integrating that
matrix into the certificate, giving the monitor more flexibility while
remaining simple.
}:

\begin{enumerate}
\item The monitor computes the logical OR of the image filtered with an edge
  detection algorithm on lightness value (such as the Sobel operator)
  and the image filtered by lightness above a certain threshold. These
  thresholds are computed by the controller and passed to the monitor
  as part of the certificate.
\item Using the bird's-eye transformation matrix passed from the
  controller, the monitor transforms the lane lines $L(i)$ and $R(i)$ from
  the bird's-eye view to the image's view.
\item For the transformed left lane curve $L_T(i)$ (and
  correspondingly for the right), the monitor creates a filter based
  on the curve, slightly blurred to allow for some margin of error in
  the proposed lane lines. The bottom of the filter is weighted more
  heavily because deviations in the region closest to the car are more
  important. At points not on the curve, the filter is padded with
  negative values so that only thin lines that match the filter's shape will produce a high
  correlation output.
\item The monitor computes the cross-correlation $C$ between the filter
  and the left side of the processed image (and correspondingly for
  the right), and finds the point of highest correlation in the image
  (namely $(i_{\max}, j_{\max})$ such that $C[i_{\max},j_{\max}] = \max(C)$). It checks
  that the point of highest correlation lies on the transformed lane
  line $L(i)$, and that the maximum correlation is above a predefined threshold.
\end{enumerate}

\subsection{Experiments}

\begin{figure}
  \centering
  \includegraphics[width=\linewidth]{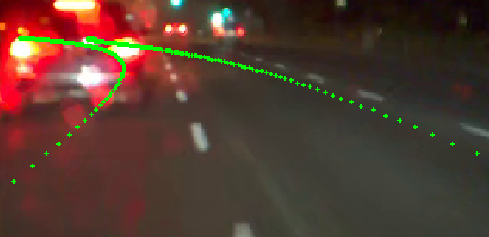}
  \caption{An example of a lane-detection failure detected by the monitor. The green lines represent the proposed lane lines.}
  \label{fig:failure}
\end{figure}

The vision certificate checking scheme was implemented and evaluated
against two different lane-detection software systems. The first was
\textit{openpilot}~\cite{CommaaiOpenpilot2020}, in which we used real replay data from a car driven using
the \textit{openpilot} system. We used the \textit{Comma2k19}
dataset~\cite{CommaAIDriving2020}, which
included driving segments with non-highway driving and adverse
lighting conditions (such as a rainy night).

In several instances, our
monitor caught lane detection failures (Fig.~\ref{fig:failure}). In all failure
cases, however, \textit{openpilot} was able to successfully correct its lane
detection within a few seconds. For some of these cases, it was
not obvious from inspection of the image taken from the camera's
perspective that the proposed lane lines were not geometrically
correct. However, viewing from the bird's-eye perspective, it was
easier to see that they could not correspond to plausible lane lines.

The certificate used for \textit{openpilot} required about 45KB of storage,
dominated by the image. On average, checking of the certificate took
about 0.18 seconds on a Dell XPS 9570 Intel Core i7-8750H CPU with
16GB RAM. This would not be adequate performance in a production
system, but it is within an order of magnitude of what would be
required. With proper optimization of the checking
algorithm, and some additional compression of the image, performance
seems unlikely to be a problem.

The second experiment was conducted with the physical racecar. We
implemented a naive lane-detection scheme, and used colored tape on
the floor to simulate lane lines. By placing additional tape segments
in inappropriate positions, we were able to get the lane detector to
report bad lane lines; in all cases, the monitor correctly rejected
them.

\subsection{Component Isolation and Authentication of Image Data}
To demonstrate security in a certified control lane following system, we augmented the visual lane detection system with authentication and timestamping of sensor data, and with component isolation. We adopted a threat model in which a malicious controller could: 

\begin{itemize}
    \item propose incorrect lane lines
    \item send a modified image or timestamp to the monitor
    \item pass stale images to the monitor
    \item not send anything at all to the monitor
\end{itemize}
For each case, we implemented a simple malicious controller that attempts to execute the associated attack. 

The system consisted of four components: the sensor, controller, monitor, and actuators. At startup, the sensor generates a private key and establishes a one-time connection with the monitor to pass its public key to the monitor. The sensor then chooses an image from a random set of dashcam images (simulating a real sensor's readings of the environment), adds a timestamp to the image, and signs it with the private key. It sends the signed image to the controller, which processes it to determine the location of the lane lines. The controller assembles its certificate (as described above) and sends that to the monitor. The monitor uses the public key from the sensor to authenticate the image. It checks that the timestamp is newer than any other image it has previously received and that it was generated within the last 0.8 seconds. It then does the computer vision checks (as previously described) to determine whether the proposed lane lines are correct. If all checks pass, the monitor sends a "continue" action to the actuator; otherwise, it sends a "stop" action. In addition, if the monitor does not receive a certificate for 0.8 seconds, it automatically sends a "stop" action to the actuator.

This scheme allowed the monitor to detect attacks from each of the implemented four malicious controllers. However, to guard against more sophisticated attacks that attempt to directly manipulate the sensor, monitor, or actuators, we also added a layer of isolation between the components. 

We utilized Docker and its OS-level virtualization to isolate each of the four components into their own Linux containers. Other stronger forms of isolation were considered, such as the use of virtual machines or hardware separation, but Docker offered the most in terms of performance and portability. To isolate communication channels between containers, we created separate bridge networks for each possible component connection. Each container was given a different IP address for each network it was a part of, e.g. the sensor had a different IP for its communication with the monitor from its communication with the controller. Both the sensor and monitor shared the same clocks as the host machine for more precise timing coordination.

\subsection{Conclusions}
The experiments with \textit{openpilot} show that lane lines detected by neural nets can be inaccurate and that these inaccuracies can be identified using classical computer vision techniques. Though these experiments were not exhaustive, they show that this technique can work on real data (from \textit{openpilot}) and when physically deployed to a model self-driving car platform. In most cases the monitor was able to correctly identify bad lane lines and verify good lane lines, which gives reason to believe that a balance of tolerance and sensitivity is possible (\textbf{RQ1}). Even though the racecar used only a naive lane-finding algorithm with no machine learning, the algorithms still used three times as many lines of code as the monitor's three vision checks combined (Fig.~\ref{fig:visioncodesizes}) (\textbf{RQ2}). Production controllers are of course much more complex---\textit{openpilot}'s deep learning container for lane-finding on Github has around 26 convolutional and 7 fully connected layers. This reduction in the size of the trusted base, coupled with the fact that the monitor uses only classical computer vision techniques (most of which are matrix convolutions), means that formal verification of the monitor, while still a tricky task, is closer within reach than verification of a machine-learning-based lane detector (\textbf{RQ3}).

\begin{figure}
  \begin{center}
    \caption{    Comparison of code sizes for vision monitor vs. certificate generation within the controller. Only the naive vision case is included; the \textit{openpilot} version uses a complex neural net.}
    \begin{tabular}{ | l | c | r |} \hline \rowcolor{Gray}
      Component & Lines of Code \\
      \hline

      Vision Monitors library calls & 90 \\
      Vision Monitors self-written code & 235 \\
      \bf{Vision Monitors Total} & \bf{325}  \\ \hline
      Controller library calls & 460 \\
      Controller self-written driver code & 612 \\
      \bf{Vision Controller Total}  & \bf{1072} \\
      \hline
    \end{tabular}
    \label{fig:visioncodesizes}
  \end{center}
\end{figure}

\subsection{Limitations}

This certificate is inherently less trustworthy than the LiDAR
certificate. This seems unavoidable, since unlike the physical
obstacles detected by the LiDAR scheme, following lane lines is a
social convention, with the lane lines acting as signs whose
interpretation is not a matter of straightforward physical properties.

\begin{figure}
  \centering
  \includegraphics[width=\linewidth]{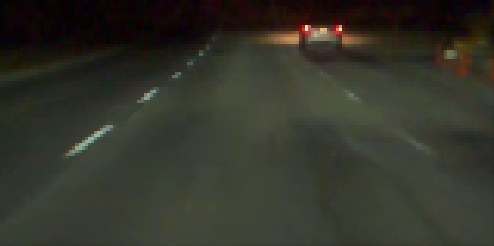}
  \caption{A frame that caused the conformance test to fail due to the faded right-hand lane line
  }
  \label{fig:bad_lines}
\end{figure}

There are two failure cases that we encountered during testing. The first involves obstructed lane
lines. Our design assumes that the lane boundaries are always at least
partially visible. Since the dataset includes segments with non-highway driving, there were cases in which
multiple cars were lined up at a stoplight and obscuring the lane lines. On the highway, the lane
lines could be similarly obstructed during traffic or when a car in front is changing lanes. The second failure
case involves poorly painted lane lines. In one case, a lane line was so faded it caused the correlation output
of the conformance test to not reach the desired threshold and therefore fail the check (Fig.~\ref{fig:bad_lines}). This situation could be
mitigated if color filter thresholds were dynamically computed and passed by the controller to the monitor. However, since we could not find such
thresholds in the \textit{openpilot} replay data, we passed constant values to the monitor. This failure case also reflects the fact that compelling
evidence of the presence of a lane will require well-drawn lane lines;
a successful deployment of autonomous cars might simply require higher
standards of road markings. It is essential to realize that certified
control does not create or even exacerbate this problem but merely
exposes it. If the designer of an autonomous vehicle were willing to
rely on inferring lanes from poorly drawn lines, the certificate
requirements could be reduced accordingly, so that the level of
confidence granted by the certificate reflect the less risk-averse
choice of the designer.

\begin{figure}
  \centering
  \includegraphics[width=\linewidth]{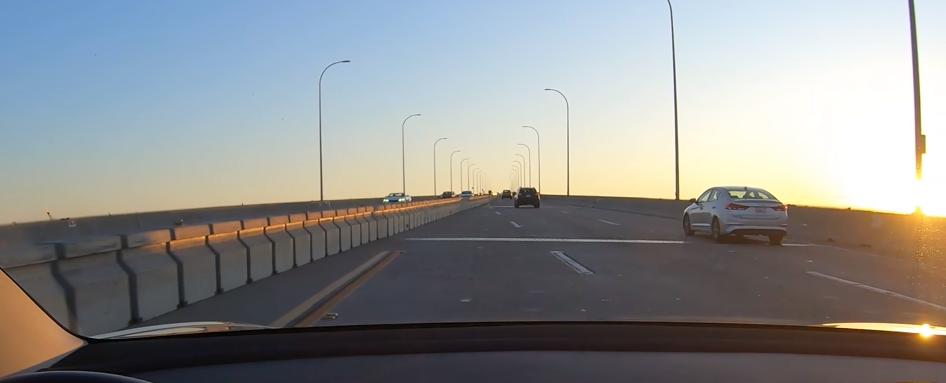}
  \caption{An image taken from dashcam footage in which a car using Tesla's autopilot almost ran into a concrete barrier. The malfunction was presumably caused by the perception system mistaking the reflection along the barrier for a lane line. }
  \label{fig:tesla}
\end{figure}

A third failure case is of particular importance because it has been
observed in erroneous behavior that has caused some cars (notably
those produced by Tesla which rely heavily on vision, since they have
no LiDAR units) to swerve towards concrete barriers. In this case, the
problem is that bright lines, that seem to be lane lines, appear in the
camera's image~\cite{TeslaAutopilotDrives}. These lines are actually bands of direct or reflected
sunlight. In most cases, we believe that our tests would catch such
anomalies; we took one particular image from an online video
(Fig.~\ref{fig:tesla}) illustrating this problem and confirmed that
the inferred lane lines would indeed fail the geometric test and be correctly rejected.

It is possible that these spurious lane lines would have passed the geometry and conformance tests. Apparently a more basic property is being violated: the detected lane line is on the barrier and not on the ground. This
motivates a new type of certificate, which we now turn to.


\section{A Certificate Combining Vision and LiDAR}

Image data lacks the inherent physical properties of LiDAR data: a
single pixel, unlike a single LiDAR point, says nothing about the
car's environment, absent other context. As a result, one can only
perform limited checks on the lane lines using vision alone. Indeed in the Tesla example, the
yellow ray of sun looked, in the image, like a plausible lane
line. However, the line was not spatially on the ground plane.

This motivated a certificate that integrates LiDAR and vision data for
verifying lane line detection. In addition to providing the lane line
polynomials and the camera image (as above), the certificate
also provides a set of LiDAR points that lie on the purported lane
lines. The monitor checks that these
LiDAR points indeed correspond to the lane lines and that they reside
on the ground plane. The controller, as usual, is given the more
complex task, in this case selecting the LiDAR points.

\subsection{Experiment}

In the controller, we take points from the image's lane lines and transform them into the
LiDAR space, yielding the corresponding LiDAR readings. Doing so,
given knowledge of the camera and LiDAR specifications and their
mounting locations on the vehicle, is a matter of basic
trigonometric transformations.

We additionally task the controller with identifying a ground
plane. Our controller implementation runs a random sample consensus
(RANSAC) algorithm~\cite{fischlerRandomSampleConsensus1981}, augmented with some constraints on necessary
ground plane features (e.g. slope, height relative to the car), to
determine the plane.

\begin{figure}
  \centering
  \includegraphics[width=\linewidth]{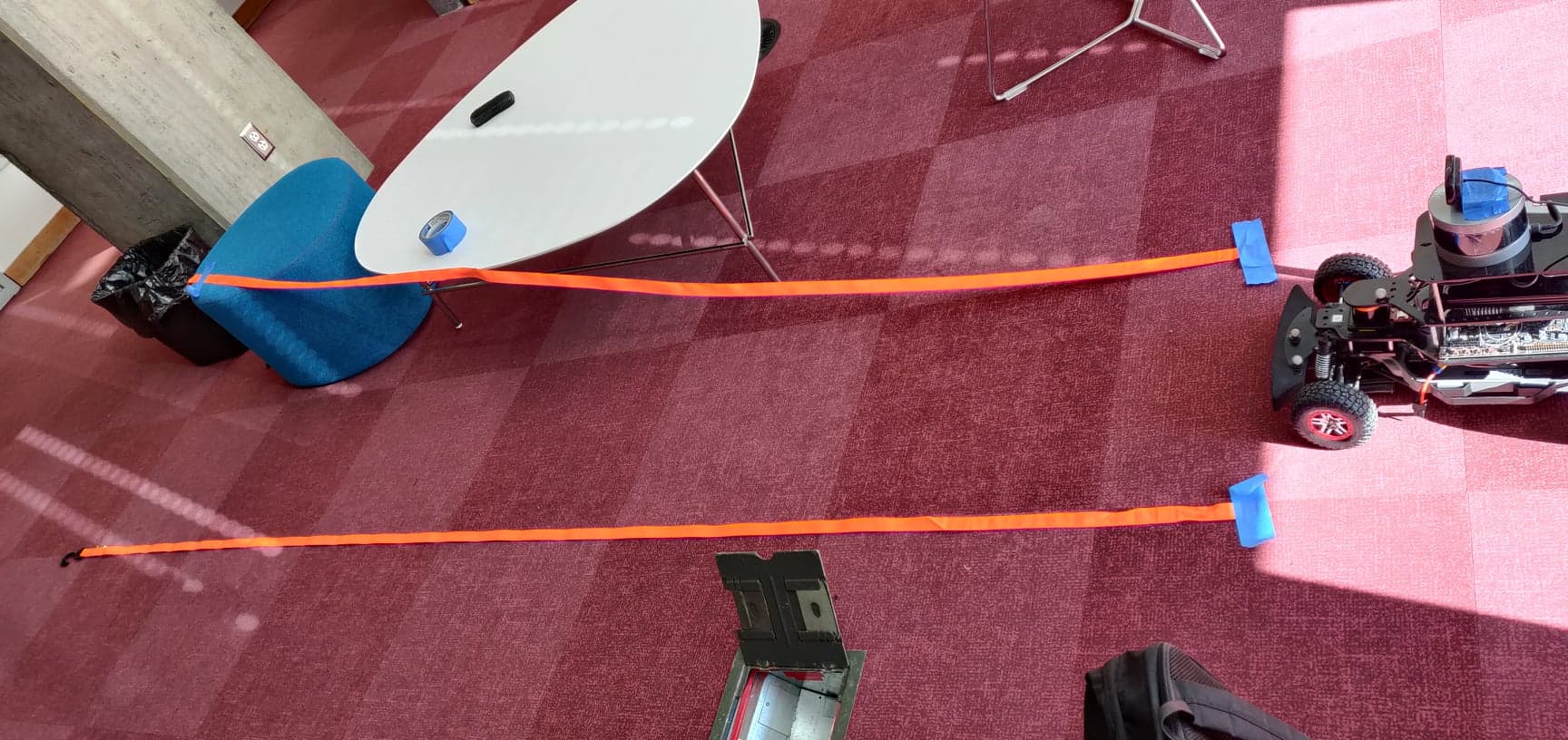}
  \caption{The experimental setup for testing a certificate that combines vision and LiDAR data, featuring a raised lane line that is visually indistinguishable (from the point of view of the car) from a normal lane line}
  \label{fig:lidarvision}
\end{figure}

Given the LiDAR lane line points and the ground plane, the monitor
checks that the lane points do lie sufficiently close to the ground
plane. To simulate the Tesla barrier malfunction, we lined up two rows of
tape to look like lane lines. The tape for the left line was placed on
the ground while the tape for the right lane was elevated on a
platform; from the camera's perspective, the pair of lane lines looked
geometrically plausible (Fig.~\ref{fig:lidarvision}). When we ran our combined vision-LiDAR
implementation on this scenario, the monitor correctly rejected the
certificate because the points from the right lane line were not on
the ground plane. This check would also detect false lane lines
which lie above or below the ground plane for any other reason.

\subsection{Evaluation}

The integration of vision with LiDAR presents a few unique
challenges. First, the ground plane detector we implemented is not
robust to steeply sloped roads or to very uneven road
surfaces.
Second, since the controller does not include in its certificate any proof of validity
of the ground plane it detected, the ground plane algorithm must be regarded as within the trusted base. The obvious remedy---namely including the ground plane detection in the monitor---is not straightforward for two reasons: the algorithm is too complex to be easily verifiable, and
the monitor only has access to the points in the certificate, not the entire LiDAR point cloud. Nevertheless, we hope to solve this problem with our usual approach: by tasking
the controller with producing a certificate which proves to the
monitor the validity of its suggested ground plane, e.g. by inclusion of certain low-lying points.

\subsection{Conclusion}
Our experiments showed that it is possible for a monitor combining LiDAR and vision data to check the detected lane lines in a reliable manner, and to catch cases that would have passed a purely visual test (\textbf{RQ1}). Compared to the complexity of a neural net employed by lane-detection software like \textit{openpilot}, being able to verify the lane lines using our monitor offers a huge reduction in the complexity of the trusted base (\textbf{RQ2}) which goes hand-in-hand with simplifying the task of formal verification (\textbf{RQ3} ).


\end{document}